\documentclass{IOS-Book-Article}

\usepackage{mathptmx}
\usepackage{soul}\setuldepth{article}
\usepackage[dvipsnames]{xcolor}
\usepackage{tcolorbox}
\usepackage{framed}

\def\hb{\hbox to 11.5 cm{}}

\begin{document}

\pagestyle{headings}
\def\thepage{}
\begin{frontmatter}              

\title{Multi-granularity Argument Mining in Legal Texts  }

\markboth{}{April 2022\hb}

\author[A, B]{Huihui Xu
\thanks{Corresponding Author: huihui.xu@pitt.edu}},
\author[A, B, C]{Kevin Ashley}

\runningauthor{B.P. Manager et al.}
\address[A]{Intelligent Systems Program, University of Pittsburgh}
\address[B]{ Learning Research and Development Center, University of Pittsburgh}
\address[C]{School of Law, University of Pittsburgh}

\begin{abstract}
In this paper, we explore legal argument mining using multiple levels of granularity. Argument mining has usually been conceptualized as a sentence classification problem. In this work, we conceptualize  argument mining as a token-level (i.e., word-level) classification problem. We  use a Longformer model to classify the tokens. Results show that token-level text classification identifies certain legal argument elements  more accurately than sentence-level text classification. Token-level classification also provides greater flexibility to analyze legal texts and to gain more insight into what the model focuses on when processing a large amount of input data.

\end{abstract}

\end{frontmatter}
\markboth{April 2022\hb}{April 2022\hb}

\section{Introduction}
Argument mining is “the automatic discovery of an argumentative text portion, and the identification of the relevant components of the argument presented there.” \cite{peldszus2013argument}. The goal is to identify and extract
the structure of inference and reasoning expressed as arguments presented in natural language \cite{lawrence2020argument}. Legal argument mining identifies and extracts arguments in legal texts. 

In previous work, we applied supervised machine learning (ML) and deep learning methods to classify sentences of legal cases in terms of the roles they play in a legal argument: the issue a court addresses, the conclusions of those issues, and the court's reasons for so concluding. We call these elements in a legal argument  IRC triples. In previous work, we demonstrated that supervised ML and deep learning methods can identify IRC types of sentences to some extent. 
  

In this paper, we take legal argument mining to a finer-grained level -- token-level argument mining where the tokens are words. That is, we treat it as a word classification task. 
Token-level argument mining has several potential advantages. First, it is more robust against errors in sentence segmentation \cite{trautmann2020fine}. Secondly, it can efficiently handle  single sentences that exhibit multiple argumentative elements. For example, as shown in Figure \ref{fig:sentence_example}, different parts of a single sentence have been labeled as conclusion and reason. If we apply sentence-level classification methods for each label to the same sentence, we confuse the classifier and  lose ordering information as compared with training on those sub-sentences.
Finally, token-level argument mining can provide insights about the contributions of particular words to  sentence-level classification. 

Our contributions in this work are, first, to apply token-level argument mining to legal texts. Secondly, we show that this token-level approach  improves the accuracy of classifying sentences in terms of legal argument elements. Finally, our error analysis shows new ways to understand the significance of certain tokens/words in classifying sentences by legal argumentative roles. 
\begin{figure}
\begin{framed}
 \textcolor{LimeGreen}{Allowing the appeal,} \textcolor{Aquamarine}{that s. 23(1) of the Social Assistance Act places a mandatory responsibility upon social service committees to provide assistance for all persons in need as defined in s. 19(e) of the Act.}
\end{framed}
\label{fig:sentence_example}
\caption{An example of a legal summary sentence whose parts are labeled with two argumentative elements. Green-colored text represents conclusion, and blue-colored text represents reasons.}
\end{figure}

\section{Related Work}
 Researchers have developed techniques to automatically identify arguments or their components in texts from a number of domains. Previous argument mining work has addressed political debate \cite{lippi2016argument}, finance \cite{chen2021opinion}, and others. Argument mining in the legal domain includes training classifiers on different types of extracted features to classify premises and conclusions \cite{moens2007automatic, mochales2007study}, investigating discursive and argumentative characteristics of legal documents \cite{mochales2008study}, identifying  argument schemes \cite{Feng2011classifying} or rhetorical roles that sentences play in legal cases \cite{saravanan2010identification} ,  summarizing legal cases in terms of argument elements \cite{xu2021toward,elaraby2022arglegalsumm}, and accounting for sentence position and embedding in legal argument classification \cite{xu2021accounting}.

Generic and legal argument mining involves three subtasks: component identification, component classification, and structure identification \cite{stab2017parsing}. Component identification focuses on separating argumentative units from non-argumentative units \cite{florou2013argument, goudas2014argument}. Component classification identifies the types of argumentative units \cite{xu2021toward, rooney2012applying}. Structure identification addresses the relations (e.g., support, attack) between argument units \cite{stab2014identifying, ghosh2014analyzing}. These sub-tasks have usually been conceptualized as involving sentence classification. For example, component identification is a binary classification task, where a sentence can either be argumentative or non-argumentative. The argumentative sentences are then classified, for instance, as premises or conclusions, and structural relations among the sentences are  identified.

Some recent argument mining research has focused on a more granular level, the token-level, which means assigning labels to every word. 
For example, \cite{trautmann2020fine} showed that the token-level argument mining employed in Argument Unit Recognition and Classification (AURC)  retrieves a larger number of arguments than  sentence-level mining. \cite{ding2022don} also treated the Kaggle competition ``Feedback Prize - Evaluating Student Writing" as a token-level argument mining task. 

Some sequential labeling techniques have been applied in this context, like BERT, Conditional Random Fields, and Bi-LSTM \cite{trautmann2020fine, ajjour2017unit}. This conceptualized token-level classification resembles the classic sequence labeling task in NLP like Named Entity Recognition (NER). Hidden Markov Models (HMM), Maximum entropy Markov models (MEMMs) \cite{freitag2000information}, and Conditional Random Fields (CRF) are the most commonly used sequential labeling techniques in the pre-neural model era. Recently, researchers have applied neural models to tackle sequence labeling problems such as convolution networks \cite{collobert2011natural}, bidirectional LSTM-CRF models \cite{huang2015bidirectional}, and BERT-CRF \cite{souza2019portuguese}. 

As far as we know, token-level argument mining has not yet been applied in legal argument mining. We have applied it to a corpus of expert-annotated legal cases and summaries as described below. 

\section{Dataset}
\subsection{IRC Sentence Annotation}
Our dataset  comprises 28,733 legal cases and summaries prepared by attorneys, members of legal societies, or law students and provided by the Canadian Legal Institute (CanLII).\footnote{\url{https://www.canlii.org/en/}}
As noted our IRC type system for labeling sentences in legal cases and case summaries includes: 
\begin{itemize}
    \item \textbf{Issue} -- Legal question which a court addressed in the case.
    \item \textbf{Conclusion} --  Court's decision for the corresponding issue.
    \item \textbf{Reason} -- Sentences that elaborate on why the court reached the Conclusion.
\end{itemize}
All un-annotated sentences are treated as non-IRC sentences.

We employed two third-year law school students  to annotate sentences from the human-prepared summaries in terms of issues, reasons, and conclusions. They annotated 1049  randomly selected case/summary pairs. 
Both annotators followed a detailed 8-page  Annotation Guide prepared by the second author, a law professor, in marking sentences in both the summaries and full texts. We first asked annotators to annotate  the summaries  since they are shorter texts prepared by human summarizers who have selected the most important information from the cases they summarized. After annotating a set of summaries, the annotators resolved any annotation differences in regular Zoom meetings by consulting an expert, the second author.

The annotators employed a different procedure for marking up the full texts. The Annotation Guide instructed them to find sentences from the full texts which are most similar to the annotated summary sentences and to assign them the same labels (i.e., issue, conclusion, or reason) as in the summaries. Annotators can use the words or phrases from the annotated summary sentences to search for the corresponding sentences in the full texts. With this approach, annotators do not need to read the full case decisions, which are much longer than summaries.  Since human summarizers tend to edit selected sentences in the full texts to make their case summaries more coherent, the full texts often lack   sentences  exactly matching  those in the annotated summary. For example, a human summarizer may combine multiple short sentences into a longer one.  In matching sentences from the summary and full text, the guidelines instructed annotators to ignore minor differences in wording or phrasing, to mark up all of the shorter sentences that appeared to have been combined into an annotated summary sentence, and to mark up all similar versions of an annotated summary sentence. By leveraging the annotated summary sentences to target the corresponding issue, reason and conclusion sentences in full texts of cases,  we gathered a sizeable annotated dataset within our time and financial constraints. 

We employed the widely-used Cohen's $\kappa$ \cite{cohen1960coefficient} metric to assess the degree of agreement between two annotators. The mean of Cohen's $\kappa$ coefficients across all types for summaries is 0.734; the mean for full texts of cases is 0.602. Both scores indicate substantial agreement between two annotators according to \cite{landis1977measurement}. The  full texts annotation agreement is lower than that of summaries since the sentences of full texts and summaries are not in a one-to-one mapping. 

Table \ref{tab:type_length} shows the statistics of our annotated summaries and full texts sentences. As shown in the table, full text sentences are longer on average than summary sentences across all types. 

\begin{table*}
  \caption{Statistics of lengths of IRC sentences in summaries and full texts. The length is measured in terms of tokens/words.}
  \label{tab:type_length}
  \setlength{\tabcolsep}{2.5pt} 
  \begin{tabular}{l|rrr|rrr}
  \hline
         &\multicolumn{3}{c|}{Summary}
         &\multicolumn{3}{c}{Full text}\\
             &\multicolumn{1}{c}{Issue}
             &\multicolumn{1}{c}{Reason}
             &\multicolumn{1}{c|}{Conclusion} 
             &\multicolumn{1}{c}{Issue}
             &\multicolumn{1}{c}{Reason}
             &\multicolumn{1}{c}{Conclusion}\\
\hline
Minimum length &2  &2  &2 &2 &2 &1 \\
Maximum length &131  &230  &269 &361 &174 &254 \\
Mean length &25  &24 &18 &31 &27 &24 \\
\hline
 \end{tabular}
\end{table*}

\subsection{Token Tagging using BIO Scheme}
\begin{figure}
    \centering
    \includegraphics[scale=0.41]{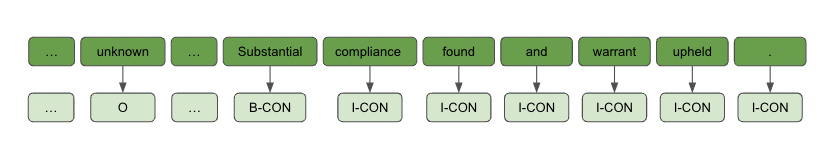}
    \caption{An example of using BIO format to tag every token in a conclusion sentence. }
    \label{fig:bio_tags}
\end{figure}

The BIO or IBO tagging scheme was first proposed in \cite{ramshaw1999text}. We adapt the BIO tagging format to our annotated summary/full text pairs. One advantage of this tagging format is it allows tokens to carry both the sentence structure and sentence type information in our work. As shown in the Figure \ref{fig:bio_tags}, the B-prefix of a tag indicates the beginning of an annotated conclusion sentence, the I-prefix of a tag indicates the token is inside  a conclusion sentence, while the O tag indicates the token does not belong to any typed sentence. 

\section{Experiment}
In this work, we first trained a model on the token-level and then used the token labeling to further determine the sentence labels. The pre-trained models we used are from Hugging Face,\footnote{\url{https://huggingface.co/}} and all programs are written in Python.

\begin{figure}
    \centering
    \includegraphics[scale = 0.5]{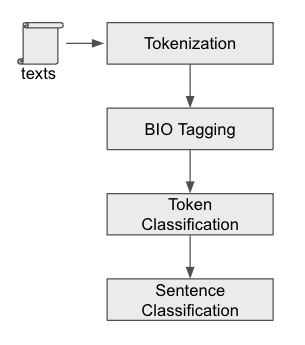}
    \caption{The experiment pipeline of this work.}
    \label{fig:pipeline}
\end{figure}

The  experiment pipeline is shown in Figure \ref{fig:pipeline}.
We pre-processed our dataset using the BIO format: we first tokenized all the sentences in summaries and full texts, then assigned the corresponding BIO tags to every  token. Those BIO-tagged tokens were then put into the pretrained Longformer \cite{Beltagy2020Longformer} model for token classification. We chose Longformer over the traditional BERT \cite{devlin2018bert} model because of its ability to process longer documents. The maximum input length is 1024 tokens due to the GPU limitation.\footnote{We use a single NVIDIA Titan X GPU with 12 GB memory.} An input length of 1024 tokens is not ideal for the full texts because of their length. Truncation and chunking are the most common methods for dealing with this limitation in practice. Truncating causes loss of information because it disregards some text. As a result, we decided to segment the full texts documents into multiple chunks of length 1024 to avoid information loss. We experimented two types of Longformer: Lonformer-base-4096 and Longformer-large-4096 \footnote{\url{https://github.com/allenai/longformer}}. 

We split our dataset into 80\% training, 10\% validation and 10\% test in summaries and full texts.

\section{Results}
Table \ref{tab:token_result} shows the results of token-level classification in summaries and full texts. As seen in the table, the classification results on the summaries are  better  than on the full texts in terms of precision, recall, and F1 score. The better results on summaries are expected because the summaries are shorter than full texts and more clearly organize the sentences. 

To determine the sentence type from the resulting token labels, we used the token type that appears most frequently in the sentence. Table \ref{tab:sentence_result} reports the results of assigning sentence type utilizing the token labels. 

For purposes of comparison, we trained three techniques on  sentence-level annotation: Legal-BERT \cite{zheng2021does}, BERT \cite{devlin2018bert} and Longformer.  Neither of these baseline techniques employ token-level annotation. We used weighted cross entropy loss when training sentences from full texts because of the more severe data imbalance issue. In order to make the results across different models comparable, we tested them on the same test set. As shown in the Table  \ref{tab:sentence_result}, Longformer(large)-BIO achieved better $F_1$ scores in sentence labeling across all sentence types (e.g., issues, reasons, and conclusions). 
\begin{table*}
  \caption{Results of BIO token-level classification on summaries and full texts. All the results are reported in terms of precision, recall and $F_1$ scores. The scores inside parentheses are produced by Longformer-base-4096, while the scores outside parentheses are produced by Longformer-large-4096.}
  \label{tab:token_result}
  \setlength{\tabcolsep}{2pt} 
  \begin{tabular}{l|rrrrrrr}
  \hline
         &\multicolumn{7}{c}{Summary}\\

             &\multicolumn{1}{c}{B-Issue}
             &\multicolumn{1}{c}{I-Issue}
             &\multicolumn{1}{c}{B-Reason}
             &\multicolumn{1}{c}{I-Reason}
             &\multicolumn{1}{c}{B-Conclusion}  
             &\multicolumn{1}{c}{I-Conclusion} 
             &\multicolumn{1}{c}{O}\\
\hline
Precision &0.83 (0.79)&0.83 (0.80)&0.72 (0.67)&0.75 (0.70)&0.83 (0.77)&0.80 (0.73)&0.78 (0.77)\\

Recall &0.79 (0.78)&0.78 (0.81)&0.80 (0.75)&0.80 (0.76)&0.84 (0.80)&0.82 (0.72)&0.75 (0.72)\\

F1-score &0.81 (0.78)&0.81 (0.81)&0.75 (0.71)&0.77 (0.73)&0.83 (0.78)&0.81 (0.72)&0.77 (0.74)\\
\hline
\hline
&\multicolumn{7}{c}{Full-texts}\\

             &\multicolumn{1}{c}{B-Issue}
             &\multicolumn{1}{c}{I-Issue}
             &\multicolumn{1}{c}{B-Reason}
             &\multicolumn{1}{c}{I-Reason}
             &\multicolumn{1}{c}{B-Conclusion}  
             &\multicolumn{1}{c}{I-Conclusion} 
             &\multicolumn{1}{c}{O}\\
\hline
Precision &0.66 (0.62)&0.80 (0.75)&0.54 (0.44)&0.69 (0.64)&0.53 (0.46)&0.65 (0.61)&0.98 (0.98)\\

Recall &0.55 (0.52)&0.70 (0.69)&0.36 (0.36)&0.63 (0.62)&0.43 (0.44)&0.63 (0.61)&0.98 (0.98)\\

F1-score &0.60 (0.56)&0.74 (0.72)&0.43 (0.40)&0.66 (0.63)&0.47 (0.45)&0.64 (0.61)&0.98 (0.98)\\
\hline
 \end{tabular}
\end{table*}

\begin{table*}
  \caption{Results of classification on summaries and full texts. All the results are reported as $F_1$ scores.   }
  \label{tab:sentence_result}
  \setlength{\tabcolsep}{2pt} 
  \begin{tabular}{l|rrrr|rrrr}
  \hline
         &\multicolumn{4}{c|}{Summary}
         &\multicolumn{4}{c}{Full text}\\
             &\multicolumn{1}{c}{Issue}
             &\multicolumn{1}{c}{Reason}
             &\multicolumn{1}{c}{Conclusion}  
             &\multicolumn{1}{c|}{Non-IRC}  

             &\multicolumn{1}{c}{Issue}
             &\multicolumn{1}{c}{Reason}
             &\multicolumn{1}{c}{Conclusion}  
             &\multicolumn{1}{c}{Non-IRC}\\

\hline
Longformer(large)-BIO &0.81&0.77&0.87&0.79&0.66&0.68&0.67&0.98\\
Longformer(base)-BIO &0.82&0.72&0.81&0.77&0.63&0.67&0.64&0.98\\
Longformer(base)-no BIO &0.75&0.73&0.80&0.75&0.49&0.30&0.49&0.95\\
Longformer(large)-no BIO &--&--&--&0.58&--&--&--&--\\
Legal-BERT &0.76&0.73&0.81&0.76&0.52&0.47&0.56&0.98  \\
BERT &0.73&0.70&0.79&0.69&0.50&0.49&0.52&0.98\\
\hline
 \end{tabular}
\end{table*}

\section{Discussion}
We trained Longformer on annotated summary and full text sentences, respectively. Due to GPU limitation, we can only train large Longformer on summaries. It confirms that the BIO approach classify the sentences more effectively. 
\subsection{Token-level Classification}
As shown in Table \ref{tab:token_result}, we can see that the $F_1$ scores of I-prefixed token types (i.e., inside) are higher than B-prefixed token types (i.e., beginning). Our intuition is that I-prefixed token types benefit from more training data, because each annotated sentence has only one beginning token while I-prefixed tokens dominate the rest of the annotated sentence. Besides, Longformer-lager-4096 performs better than Longformer-base-4096 across all types. We use the best model for the following discussion.

\begin{figure}
    \centering
    \includegraphics[scale= 0.4]{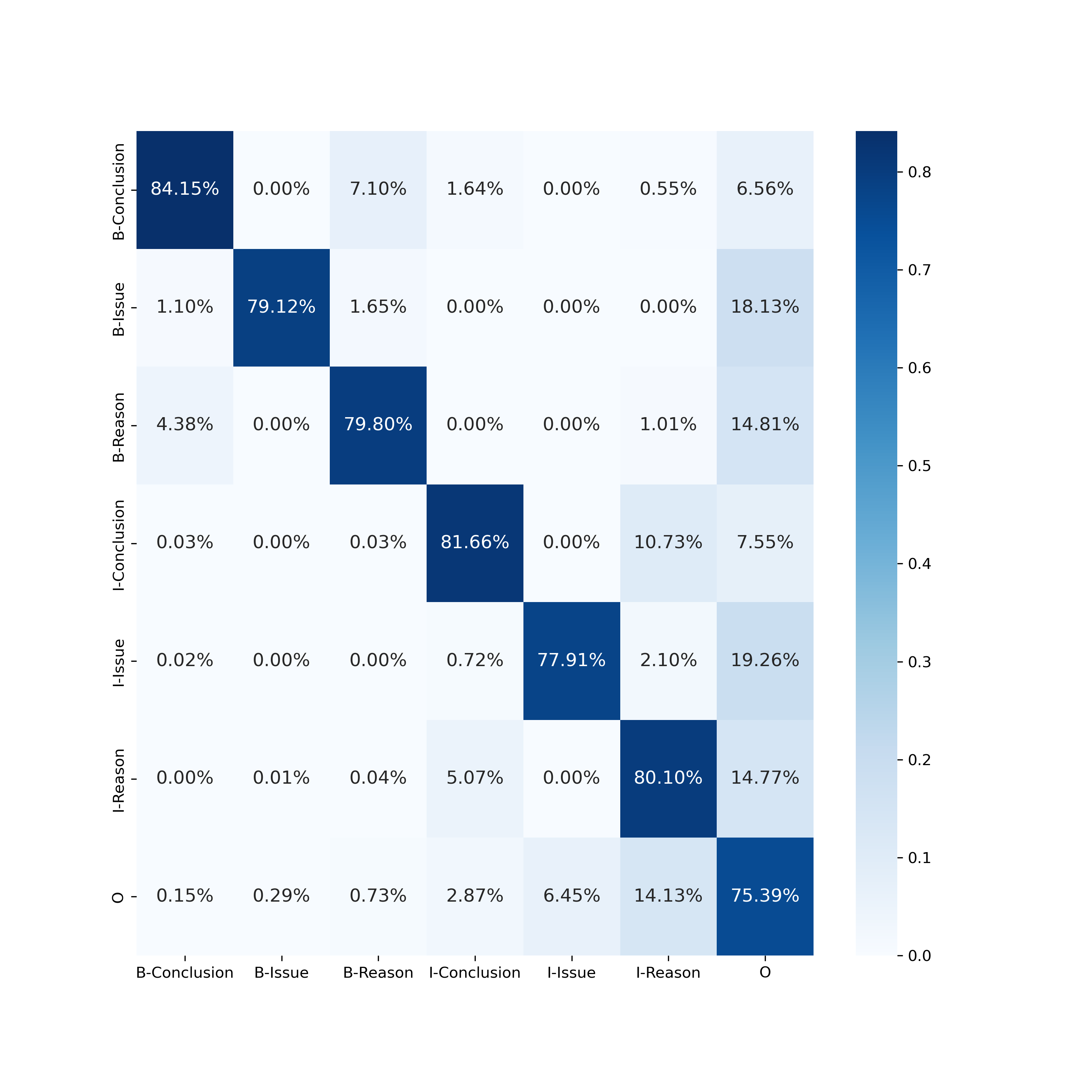}
    \caption{The confusion matrix of token-level classification in summaries.}
    \label{fig:token_confusion_summ}
\end{figure}

\begin{figure}
    \centering
    \includegraphics[scale=0.4]{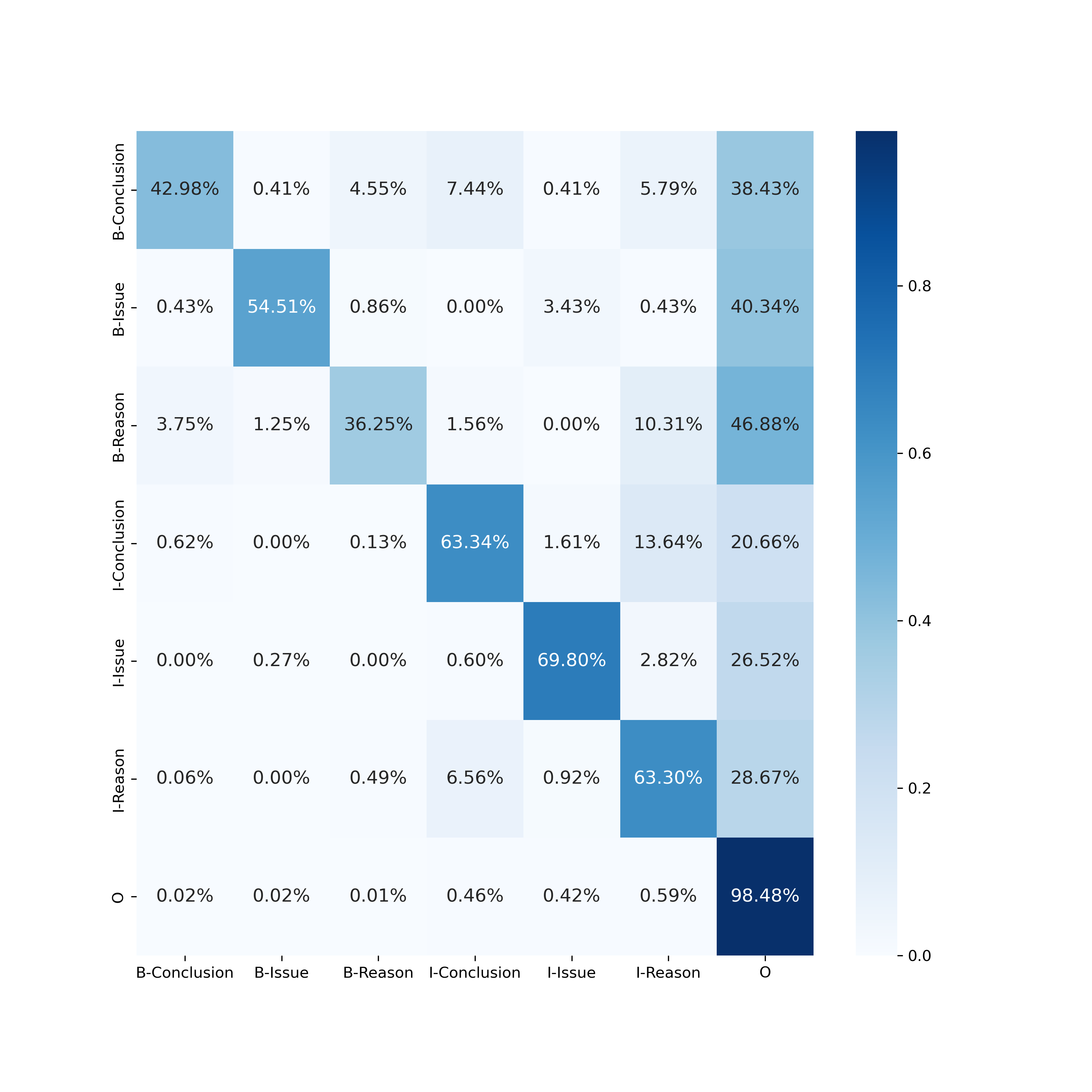}
    \caption{The confusion matrix of token-level classification in full texts.}
    \label{fig:token_confusion_full}
\end{figure}

The confusion matrices are shown for token-level classification in summaries (Figure \ref{fig:token_confusion_summ}) and in full texts (Figure \ref{fig:token_confusion_full}). The numbers are percentages of model-predicted labels with respect to the corresponding total number of ground truth labels. The Y axis represents the true labels; the X axis represents the predicted labels. We find that the model is more likely to assign I-Reason to a Non-IRC (O) type in both summaries and full texts. A large portion of those misclassified tokens are stop words, like `the', `to', `of' etc., which are commonly used within issue sentences. Those stop words, of course, appear everywhere in a document;  their type depends more on their context than their semantic meaning. The token-based classification indicates some contexts where even a stop word like `the' appears to have an effect.

As shown in the figures, the diagonal represents the percentages of tokens that are  classified correctly. The model correctly classified 84.15\% of the B-Conclusion tokens in summaries and 69.80\% of the I-Issue tokens in full texts. We  found that `HELD' appears  most frequently in the correctly classified B-Conclusion tokens in the summaries; `the' appears most frequently in the correctly classified I-Issue tokens in the full texts. The human summarizers tend to make conclusions more noticeable to readers by using indicators such as `HELD'. Those indicators are captured by the model. 

\subsection{Sentence-level Classification}
\begin{figure}
    \centering
    \includegraphics[scale=0.5]{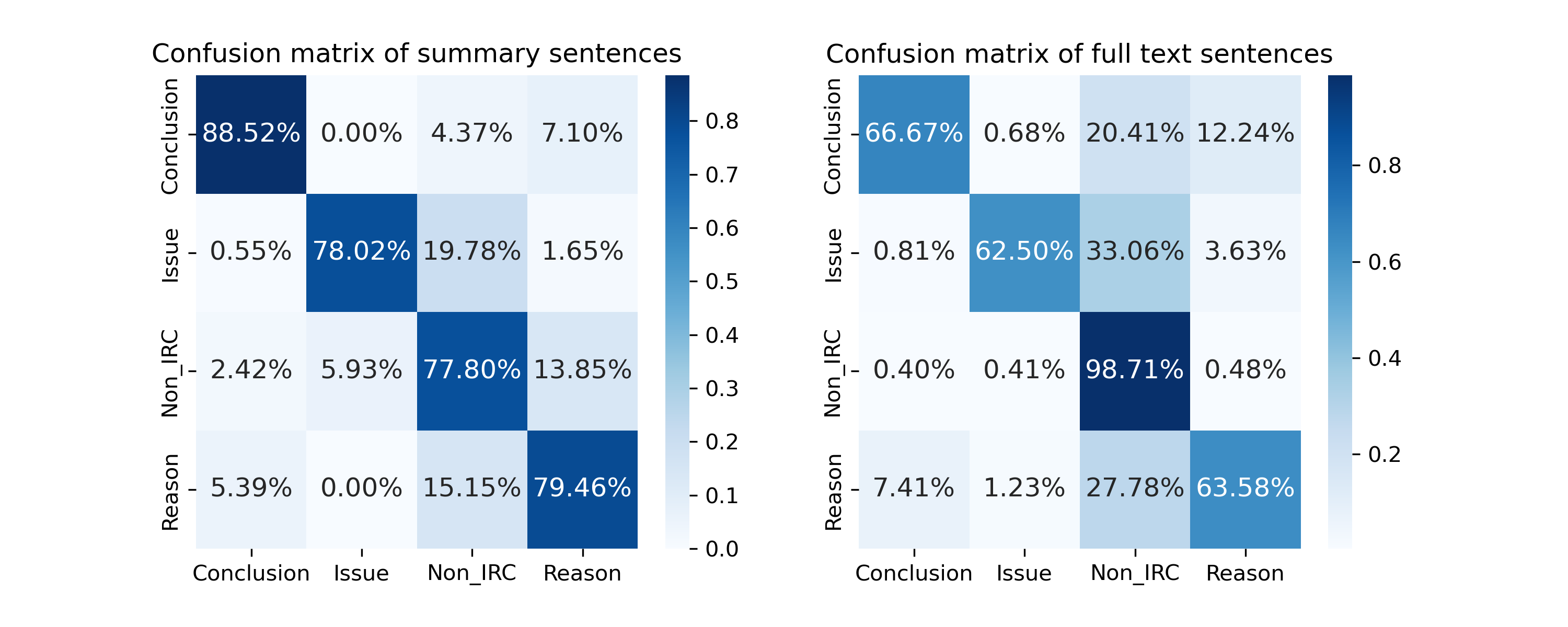}
    \caption{Confusion matrices of sentence-level classification in summaries and full texts. }
    \label{fig:sentence_confusion}
\end{figure}

As noted, the sentence type is determined by the token type that appears  most often in the sentence. Figure \ref{fig:sentence_confusion} shows the confusion matrices of sentence classifications in summaries and full texts. According to the confusion matrices, misclassified IRC sentences are most likely to be misclassified as Non-IRC  sentences in both summaries and full texts. We observed, however, that conclusions in summaries are prone to be misclassified as reasons. We investigated those misclassified conclusion sentences and find most sentences were completely misclassifed  on a token-level. That is, the model identified no conclusion tokens. Only one sentence had several correctly identified conclusion tokens including `support' and `allowed'. We have been unable to explain the token-level misclassification.

\section{Conclusion}
In this work, we experimented with multi-granular argument mining from legal texts. We employed two label classification tasks: token-level (i.e., word-level) classification and sentence-level classification. The sentence-level classification is based on the results of the token-level classification. Results showed that token-level classification  achieved more accurate sentence classification than state-of-the-art sentence-classification models. The token-level classification not only improved the sentence classification performance but also gave insights into how the model behaves with respect to certain tokens. 


In future work, we plan to use the token-based approach to more accurately classify issues, conclusions, and reasons and to use these IRC argument elements to improve automatic case summarization. We will explore  using these finer-grained indicators to identify other legal argumentative units, such as factors, and to better evaluate the quality of legal summaries in terms of coverage of argument elements.

\section*{Acknowledgement}

This work has been supported by grants from the Autonomy through Cyberjustice Technologies Research Partnership at the University of Montreal Cyberjustice Laboratory and the National Science Foundation, grant no. 2040490, FAI: Using AI to Increase Fairness by Improving Access to Justice. The Canadian Legal Information Institute provided the corpus of paired legal cases and summaries.
This work was supported in part 
by the University of Pittsburgh Center for Research Computing through the     
resources provided. Specifically, this work used the H2P cluster, which is    supported by NSF award number OAC-2117681. 
\bibliographystyle{vancouver}
\bibliography{ios-bibliography.bib}

\begin{thebibliography}{10}

\bibitem{peldszus2013argument}
Peldszus A, Stede M.
\newblock From argument diagrams to argumentation mining in texts: A survey.
\newblock International Journal of Cognitive Informatics and Natural
  Intelligence (IJCINI). 2013;7(1):1-31.

\bibitem{lawrence2020argument}
Lawrence J, Reed C.
\newblock Argument mining: A survey.
\newblock Computational Linguistics. 2020;45(4):765-818.

\bibitem{trautmann2020fine}
Trautmann D, Daxenberger J, Stab C, Sch{\"u}tze H, Gurevych I.
\newblock Fine-grained argument unit recognition and classification.
\newblock In: Proceedings of the AAAI Conference on Artificial Intelligence.
  vol.~34; 2020. p. 9048-56.

\bibitem{lippi2016argument}
Lippi M, Torroni P.
\newblock Argument mining from speech: Detecting claims in political debates.
\newblock In: Proceedings of the AAAI Conference on Artificial Intelligence.
  vol.~30; 2016. .

\bibitem{chen2021opinion}
Chen CC, Huang HH, Chen HH.
\newblock From opinion mining to financial argument mining.
\newblock Springer Nature; 2021.

\bibitem{moens2007automatic}
Moens MF, Boiy E, Palau RM, Reed C.
\newblock Automatic detection of arguments in legal texts.
\newblock In: Proceedings of the 11th international conference on Artificial
  intelligence and law; 2007. p. 225-30.

\bibitem{mochales2007study}
Mochales-Palau R, Moens M.
\newblock Study on sentence relations in the automatic detection of
  argumentation in legal cases.
\newblock Frontiers in Artificial Intelligence and Applications. 2007;165:89.

\bibitem{mochales2008study}
Mochales R, Moens MF.
\newblock Study on the structure of argumentation in case law.
\newblock In: Proceedings of the 2008 conference on legal knowledge and
  information systems; 2008. p. 11-20.

\bibitem{Feng2011classifying}
Feng VW, Hirst G.
\newblock Classifying arguments by scheme.
\newblock In: Proceedings of the 49th annual meeting of the association for
  computational linguistics: Human language technologies; 2011. p. 987-96.

\bibitem{saravanan2010identification}
Saravanan M, Ravindran B.
\newblock Identification of rhetorical roles for segmentation and summarization
  of a legal judgment.
\newblock Artificial Intelligence and Law. 2010;18(1):45-76.

\bibitem{xu2021toward}
Xu H, Savelka J, Ashley KD.
\newblock Toward summarizing case decisions via extracting argument issues,
  reasons, and conclusions.
\newblock In: Proceedings of the eighteenth international conference on
  artificial intelligence and law; 2021. p. 250-4.

\bibitem{elaraby2022arglegalsumm}
Elaraby M, Litman D.
\newblock ArgLegalSumm: Improving Abstractive Summarization of Legal Documents
  with Argument Mining.
\newblock arXiv preprint arXiv:220901650. 2022.

\bibitem{xu2021accounting}
Xu H, Savelka J, Ashley KD.
\newblock Accounting for sentence position and legal domain sentence embedding
  in learning to classify case sentences.
\newblock In: Legal Knowledge and Information Systems. IOS Press; 2021. p.
  33-42.

\bibitem{stab2017parsing}
Stab C, Gurevych I.
\newblock Parsing argumentation structures in persuasive essays.
\newblock Computational Linguistics. 2017;43(3):619-59.

\bibitem{florou2013argument}
Florou E, Konstantopoulos S, Koukourikos A, Karampiperis P.
\newblock Argument extraction for supporting public policy formulation.
\newblock In: Proceedings of the 7th Workshop on language technology for
  cultural heritage, social sciences, and humanities; 2013. p. 49-54.

\bibitem{goudas2014argument}
Goudas T, Louizos C, Petasis G, Karkaletsis V.
\newblock Argument extraction from news, blogs, and social media.
\newblock In: Hellenic Conference on Artificial Intelligence. Springer; 2014.
  p. 287-99.

\bibitem{rooney2012applying}
Rooney N, Wang H, Browne F.
\newblock Applying kernel methods to argumentation mining.
\newblock In: Twenty-Fifth International FLAIRS Conference; 2012. .

\bibitem{stab2014identifying}
Stab C, Gurevych I.
\newblock Identifying argumentative discourse structures in persuasive essays.
\newblock In: Proceedings of the 2014 conference on empirical methods in
  natural language processing (EMNLP); 2014. p. 46-56.

\bibitem{ghosh2014analyzing}
Ghosh D, Muresan S, Wacholder N, Aakhus M, Mitsui M.
\newblock Analyzing argumentative discourse units in online interactions.
\newblock In: Proceedings of the first workshop on argumentation mining; 2014.
  p. 39-48.

\bibitem{ding2022don}
Ding Y, Bexte M, Horbach A.
\newblock Don’t Drop the Topic-The Role of the Prompt in Argument
  Identification in Student Writing.
\newblock In: Proceedings of the 17th Workshop on Innovative Use of NLP for
  Building Educational Applications (BEA 2022); 2022. p. 124-33.

\bibitem{ajjour2017unit}
Ajjour Y, Chen WF, Kiesel J, Wachsmuth H, Stein B.
\newblock Unit segmentation of argumentative texts.
\newblock In: Proceedings of the 4th Workshop on Argument Mining; 2017. p.
  118-28.

\bibitem{freitag2000information}
Freitag D, McCallum A.
\newblock Information extraction with HMM structures learned by stochastic
  optimization.
\newblock AAAI/IAAI. 2000;2000:584-9.

\bibitem{collobert2011natural}
Collobert R, Weston J, Bottou L, Karlen M, Kavukcuoglu K, Kuksa P.
\newblock Natural language processing (almost) from scratch.
\newblock Journal of machine learning research. 2011;12(ARTICLE):2493-537.

\bibitem{huang2015bidirectional}
Huang Z, Xu W, Yu K.
\newblock Bidirectional LSTM-CRF models for sequence tagging.
\newblock arXiv preprint arXiv:150801991. 2015.

\bibitem{souza2019portuguese}
Souza F, Nogueira R, Lotufo R.
\newblock Portuguese named entity recognition using BERT-CRF.
\newblock arXiv preprint arXiv:190910649. 2019.

\bibitem{cohen1960coefficient}
Cohen J.
\newblock A coefficient of agreement for nominal scales.
\newblock Educational and psychological measurement. 1960;20(1):37-46.

\bibitem{landis1977measurement}
Landis JR, Koch GG.
\newblock The measurement of observer agreement for categorical data.
\newblock biometrics. 1977:159-74.

\bibitem{ramshaw1999text}
Ramshaw LA, Marcus MP.
\newblock Text chunking using transformation-based learning.
\newblock In: Natural language processing using very large corpora. Springer;
  1999. p. 157-76.

\bibitem{Beltagy2020Longformer}
Beltagy I, Peters ME, Cohan A.
\newblock Longformer: The Long-Document Transformer.
\newblock arXiv:200405150. 2020.

\bibitem{devlin2018bert}
Devlin J, Chang MW, Lee K, Toutanova K.
\newblock Bert: Pre-training of deep bidirectional transformers for language
  understanding.
\newblock arXiv preprint arXiv:181004805. 2018.

\bibitem{zheng2021does}
Zheng L, Guha N, Anderson BR, Henderson P, Ho DE.
\newblock When Does Pretraining Help? Assessing Self-Supervised Learning for
  Law and the CaseHOLD Dataset.
\newblock arXiv preprint arXiv:210408671. 2021.

\end{thebibliography}

\end{document}